\pdfoutput=1

\documentclass[11pt]{article}

\usepackage[final]{acl}

\usepackage{times}
\usepackage{latexsym}
\usepackage{placeins}
\usepackage[T1]{fontenc}

\usepackage[utf8]{inputenc}

\usepackage{microtype}

\usepackage{inconsolata}

\usepackage{graphicx}
\usepackage{pdflscape}  
\usepackage{array}       
\usepackage{adjustbox}
\usepackage{booktabs}    
\usepackage{multirow}    
\usepackage{ragged2e} 
\usepackage{xcolor}
\usepackage{cellspace} 
\usepackage{amsmath}
\usepackage{algorithmicx}
\usepackage{algpseudocode}
\usepackage{booktabs}
\usepackage{listings}

\usepackage{enumitem}

\usepackage{tikz}

\usepackage[font=footnotesize,labelformat=simple]{subcaption}
\usepackage[T1]{fontenc}

\title{Competing LLM Agents in a Non-Cooperative Game of Opinion Polarisation}

\author{
 \textbf{Amin Qasmi\textsuperscript{2}},
 \textbf{Usman Naseem\textsuperscript{3}},
 \textbf{Mehwish Nasim\textsuperscript{1}}
 \\
 \textsuperscript{1}The University of Western Australia,
 \\
 \textsuperscript{2}Lahore University of Management Sciences,
 \\
 \textsuperscript{3}Macquarie University,
\\
 \small{
   \textbf{Correspondence:} \href{mailto:mehwish.nasim@uwa.edu.au}{mehwish.nasim@uwa.edu.au}
 }
}

\begin{document}
\maketitle
\begin{abstract}



We introduce a novel non-cooperative game to analyse opinion formation and resistance, incorporating principles from social psychology such as confirmation bias, resource constraints, and influence penalties.  Our simulation features Large Language Model (LLM) agents competing to influence a population, with penalties imposed for generating messages that propagate or counter misinformation. This framework integrates resource optimisation into the agents' decision-making process.  Our findings demonstrate that while higher confirmation bias strengthens opinion alignment within groups, it also exacerbates overall polarisation. Conversely, lower confirmation bias leads to fragmented opinions and limited shifts in individual beliefs.  Investing heavily in a high-resource debunking strategy can initially align the population with the debunking agent, but risks rapid resource depletion and diminished long-term influence.


\end{abstract}

\section{Introduction and Background}

The study of opinion dynamics, originating from efforts to understand how individuals modify their views under social influence \citep{kelman1958compliance, kelman1961american}, has broad applications in areas such as public health campaigns, conflict resolution, and combating misinformation.  Within social networks, opinions spread and evolve, influenced by various factors including peer interactions \cite{kandel1986processes}, media exposure \cite{zucker1978variable}, and group dynamics \cite{friedkin2011social}. Developing accurate models of these processes is essential not only for predicting trends like opinion polarisation \citep{Small2024} or consensus formation but also for crafting targeted interventions to mitigate harmful effects, such as the spread of misinformation or societal fragmentation \citep{hegselmann2015opinion}. Agent-based models (ABMs) simulate interactions among individual agents to examine the emergent properties of opinion dynamics.
These models offer robust frameworks for analysing complex scenarios \cite{deffuant2002can, mathias2016bounded}, evaluating strategies to reduce negative consequences, and potentially fostering constructive social influence by integrating explicit cognitive mechanisms into opinion-updating processes.

This work explores the potential of Large Language Models (LLMs) to simulate human-like opinion dynamics and influence propagation within social networks. Traditional agent-based models (ABM) rely on carefully designed rules to approximate the complexity of human communicative behaviours. While these rules can be realistic, the agents themselves lack the cognitive and linguistic depth of humans, raising questions about the reliability and ecological validity of the resulting simulations. The emergence of LLMs enables the use of agents that act as closer proxies for human reasoning and discourse. Building on this capability, we propose a novel non-cooperative game framework in which adversarial LLM agents, one disseminating misinformation and the other countering it, interact within a simulated population.


Unlike previous studies that model passive opinion evolution \cite{wang2025decoding, chuang2024simulating}, this work introduces a non-cooperative game in which LLM agents engage in adversarial interactions, simulating the strategic spread and counteracting of misinformation. While previous research has focused on social media structures and mitigation via nudging, our model emphasises resource-constrained influence operations and analyses the effectiveness of debunking strategies in competitive environments.

We pose the following research questions:

\begin{enumerate} [label=\textbf{RQ}\arabic*, nosep]
    \item What are the emergent behaviors in networks of agents influenced by competing LLMs?
    \item How does the competition between LLM agents shape the evolution of opinion clusters over time, also known as echo-chambers? 
    
  
\end{enumerate}

    


\begin{figure*}[t!]
    \centering
        \includegraphics[width=0.75\textwidth]{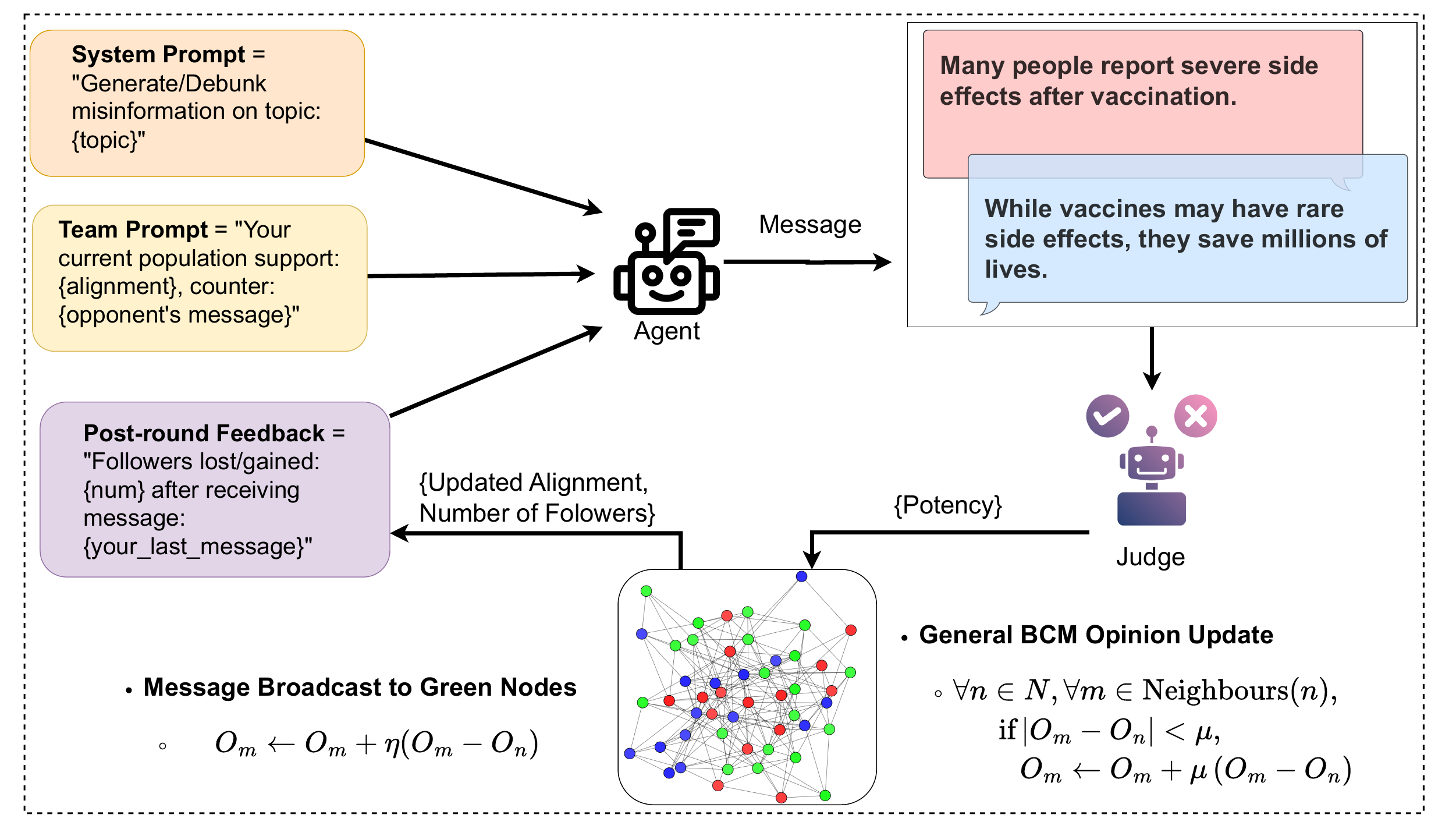}
    \caption{In each round, the active team (Red/Blue) generates a message that receives a potency value from the judge. The network updates according to the BCM algorithm. In the next round, the opposing team receives the potency results of their rival's message and their own from the previous round.}
    \label{fig:pipeline}
\end{figure*}
\section{Methodology}

We start by defining key terms used in our setup:

\begin{itemize}
    \item \textbf{Energy}: A numeric representation of a debunking agent's available communication resources. The debunking agent begins with a finite energy budget, which is depleted as they compose and send messages.

    \item \textbf{Potency}: A scalar score assigned to a message by a third-party judge agent, reflecting how persuasive or impactful the message is. Potency influences both the cost of sending a message and the likelihood of shifting another agent's opinion.

    \item \textbf{Influence Factor}: A multiplier that scales how much a message's potency can affect the opinion of a recipient. It controls the degree of opinion shift in the Bounded Confidence Model.
\end{itemize}

We use LLMs to simulate the propagation and debunking of misinformation on social media within a non-cooperative game framework.


\noindent\textbf{Scenario:}  
Our scenario models an asymmetric information environment, highlighting the challenges faced by the "Blue Team" (countering misinformation). This setup reflects adversarial dynamics commonly seen in serious games or wargames \cite{InformationWarfighter}. The "Red Team" and "Blue Team" construct, familiar in cybersecurity practices, is adapted from NIST's Glossary \cite{NISTGlossaryRedTeam}. The system comprises two LLM-based agents: the \emph{Red Agent} spreads misinformation, while the \emph{Blue Agent} debunks it. These agents operate within a directed network of neutral agents, termed \emph{Green Nodes}, representing individuals in a population. Figure \ref{fig:pipeline} illustrates this non-cooperative game structure.

\noindent\textbf{Agent Roles and Mechanics:} The simulation incorporates the following agent roles:

The \textbf{Red Agent} aims to amplify doubt and confusion by generating misinformation of varying potency.  Messages of higher potency incur penalties through rejection, reflecting real-world scenarios where informed populations are sceptical of, and less susceptible to, high-strength misinformation.



The \textbf{Blue Agent} counters misinformation from the Red Agent while operating under a resource constraint. 
The cost of generating a counter-message in round $t$ is defined as:  

\begin{equation} \label{eq:cost}
C_t = \frac{E_t}{100} \cdot P_t \cdot C_{\max}
\end{equation}

where:  
\begin{itemize}
    \item $C_t$ = cost of the counter-message at round $t$,
    \item $E_t$ = remaining energy (scaled 0--100) at round $t$,
    \item $P_t$ = potency of the message (unitless scalar),
    \item $C_{\max}$ = maximum per-message cost (fixed at $5$).
\end{itemize}

This simple linear-decay cost function reflects the intuition that generating high-potency messages consumes more energy, especially when the agent still has a large remaining energy reserve. As the agent’s energy depletes, the same potency requires less cost, encouraging more conservative behavior in later rounds. The choice of  $C_{\max} = 5$  was made to ensure that agents remain active across the full 100 simulation rounds, avoiding early exhaustion. While this formulation is straightforward, it is sufficient for modeling energy-aware communication tradeoffs and allows for future exploration of more complex cost dynamics.

To assess the impact of the energy constraint, we conducted baseline experiments in which the \textbf{Blue Agent} operated without any resource limitations. These runs replicated all experimental conditions but removed the cost equation and all references to energy constraint from the judge's and blue agent's prompts. This setup enables a direct comparison to evaluate how energy-aware behavior influences polarisation and opinion dynamics.


\textbf{Judge Agent:} This agent enables a consistent external measure of message strength, decoupled from sender or receiver bias. Each message is evaluated based on certain criteria, including \emph{clarity}, \emph{evidence}, \emph{logical reasoning}, \emph{relevance}, and \emph{impact}. The Judge and Blue agents' prompts are available in Appendix \ref{sec:prompts}.

\noindent\textbf{Simulation settings:}  The simulation begins with $n$ nodes, of which $x$ are initially aligned with the Red Agent's misinformation (pro-conspiracy), $y < n - x$ are aligned against it (anti-conspiracy), and the remaining $z = n - x - y$ are neutral. Both Red and Blue Agents generate messages, the potencies of which are determined by the Judge Agent. 


\noindent\textbf{Opinion Modeling:} We use the Bounded Confidence Model (BCM) \cite{mathias2016bounded} to simulate opinion dynamics.  In the BCM, a node updates its opinion if the difference between its opinion and that of a neighbouring node is less than a threshold (the confirmation bias value, $\mu$). The opinion update condition and formula are summarised below:


\label{bcmalgorithm}
\begin{algorithmic}
\ForAll{$n \in N$}
    \ForAll{$m \in \text{Neighbours}(n)$}
        \If{$| O_m - O_n | < \mu$}
            \State $O_m \gets O_m + \mu (O_m - O_n)$
        \EndIf
    \EndFor
\EndFor
\end{algorithmic}


where $N$ is the set of all nodes, $O_n$ is the opinion of node $n$, and $O_m$ is the opinion of its neighbour $m$. Opinions range between [-1, 1]. Nodes with opinions less than -0.5 are considered aligned with the Blue Agent, those greater than 0.5 with the Red Agent, and those in between are neutral. Nodes were initialised with a random opinion value within these thresholds.

While our goal is to explore LLM-based agent behaviors, we use BCM as a lightweight update mechanism to ensure interpretability and tractability in initial simulations.

Our model simulates opinion dynamics using a two-step process for each round of interaction. First, a message is generated by either the Red or Blue team, characterized by a specified \textit{potency} that determines the strength of the message, and an \textit{influence factor} that scales its impact on the Green network nodes. The message is broadcast to all green (neutral) nodes in the network. Each green node updates its opinion based on the following update rule, provided the BCM threshold condition is met: $O_m \gets O_m + \eta (O_m - O_n)$, where $\eta$ is a scaling factor that is proportional to the message potency, calculated as potency multiplied by the influence factor (both values ranging from 0 to 1), while $m$ is a neutral node adjacent to a blue or red neighboring node $n$.

Following this broadcast, a general network interaction is triggered in which all nodes in the network influence their neighbors using the opinion update algorithm provided earlier in this section.


\noindent\textbf{Topics Classification:} Topics for misinformation include serious debates and popular conspiracy theories (e.g., "The Earth is Flat") as well as more frivolous claims (e.g., "The Moon is made of cake"). All experimental conditions were run on 10 topics, the list of which is given in Appendix \ref{sec:topics}. The question as to whether a topic should be considered serious or satirical has been left open-ended for readers to decide.


\textbf{Models:} Our study employs GPT-4O and 4O-MINI as judges \cite{hurst2024gpt, openai2024gpt4omini}. \emph{Experiment A} compares Mixtral-8x7B-Instruct with Gemma-2-9b \cite{jiang2024mixtral, team2024gemma}, while \emph{experiment B} evaluates Mixtral-8x7B-Instruct against Gemini 1.5 Flash-8b \cite{team2024gemini}. Lastly, \emph{experiment C} contrasts Gemma-2-9b with Gemini 1.5 Flash-8b.


\noindent\textbf{Post Round Feedback:} In our simulations, agents receive feedback after each round, including their last message and metrics such as the percentage of followers gained or lost. This feedback allows them to refine their messaging strategies.


\section{Simulations and Evaluation}

We ran 100-round simulations using a directed small-world network of 50 nodes, with 40\% (20 nodes) initially aligned with the Blue Agent (anti-conspiracy) and 20\% (10 nodes) with the Red Agent (pro-conspiracy), reflecting the minority status of conspiracy theorists in social media populations \cite{Gundersen2023, ROCHERT2022101866}. The Blue Agent started with a resource value of 100 and an influence factor of 0.6, while the Red Agent's influence factor was 0.5. We tested three BCM thresholds ($\mu$): 0.3, 0.7, and 0.9.


To assess the impact of increased resource investment in debunking, we ran additional simulations (\(\mu = 0.9\)) where the Blue Agent delivered high-potency messages in the first 20 rounds (10 messages per agent). These messages had their base potency scaled by 1.2, capped at 100\% (i.e., \(\min(\text{potency} \times 1.2, 1.0)\)), simulating a "high-resource" debunking strategy. 

Simulations were performed on a local machine with an Intel i7-1355U (13th Gen) CPU and 32 GB RAM, using an integrated Intel Iris Xe GPU (15.8 GB shared memory) without dedicated GPU acceleration. For all LLMs, settings were: temperature = 0.5, top\_p = 1.0, and max\_tokens = 100. Results are presented in Section~\ref{results}.

\noindent\textbf{Metrics:} We evaluate our simulations using the following metrics:

\noindent\textbf{Polarisation} quantifies the extent of division into opposing factions within a network, reinforcing extreme views. It is calculated as follows \cite{chitra2020analyzing}:
\begin{equation} \label{polarisation}
P = \frac{1}{N} \sum_{n \in \mathcal{V}} (O_n - \bar{O})^2
\end{equation}

where $N$ is the total number of nodes, $\mathcal{V}$ is a list of all nodes, $O_n$ is the opinion of node $n$, and $\bar{O}$ is the average opinion.



\begin{table}[h]
    \centering
    \resizebox{0.5\textwidth}{!}{
    \begin{tabular}{ccccc}
        \toprule
        \textbf{BCM Threshold} & \textbf{Experiment} & \textbf{ICC Value} & \textbf{Krippendorff's Alpha} \\
        \midrule
        \multirow{3}{*}{0.3} & A & 0.660 & 0.653 \\
                             & B & 0.711 & 0.702 \\
                             & C & 0.707 & 0.702 \\
        \midrule
        \multirow{3}{*}{0.7} & A & 0.697 & 0.689 \\
                             & B & 0.780 & 0.776 \\
                             & C & 0.726 & 0.721 \\
        \midrule
        \multirow{3}{*}{0.9} & A & 0.581 & 0.567 \\
                             & B & 0.755 & 0.751 \\
                             & C & 0.710 & 0.706 \\
        \bottomrule
    \end{tabular}
    }
    \caption{Across Topics Average Intraclass Correlations (ICC) and Krippendorff's Alpha.}
    \label{tab:icc_krippendorff_values}
\end{table}

\captionsetup[subfigure]{skip=0.5pt} 
\begin{figure*}[ht]

    \subfloat[\label{fig:opinions}]{\includegraphics[width=0.7\textwidth]{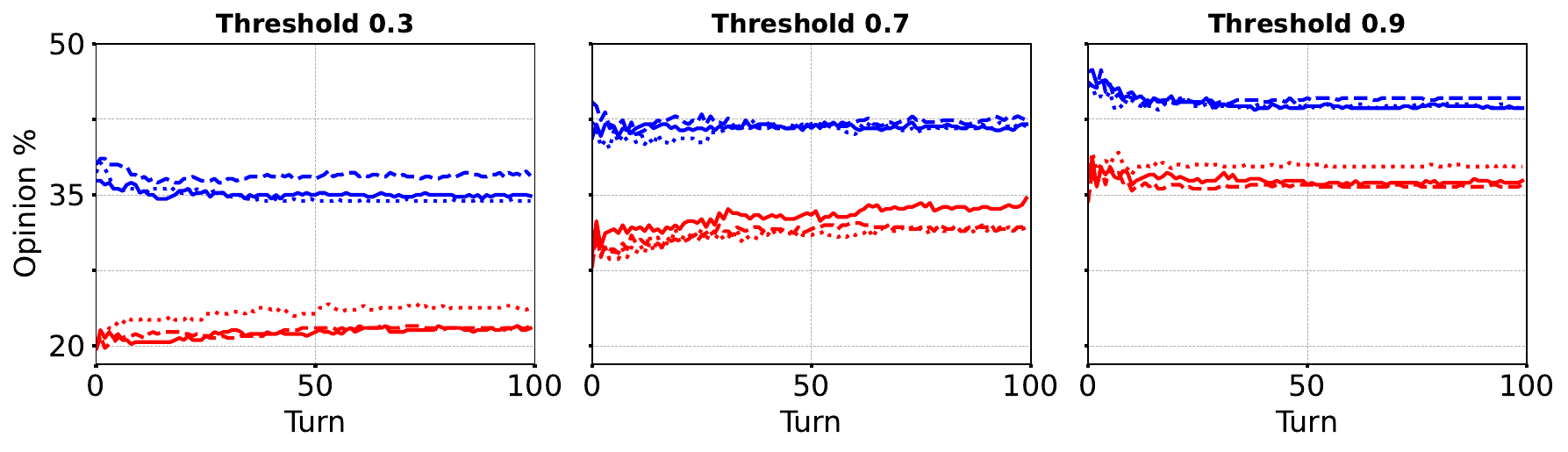}}
    \hfill 
\tikz{\draw[-,black, densely dashed, thick](0,-2.05) -- (0,1.05);} 
    \vspace{-6mm}
    \subfloat[\label{fig:opinions2}]{
        \includegraphics[width=0.275\textwidth]{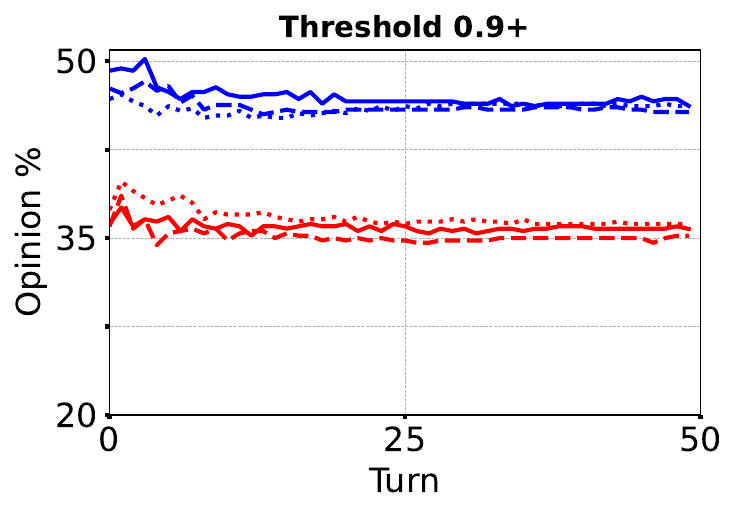} }
         \\ 
   
    \subfloat[\label{fig:polarisations}]{
        \includegraphics[width=0.68\textwidth]{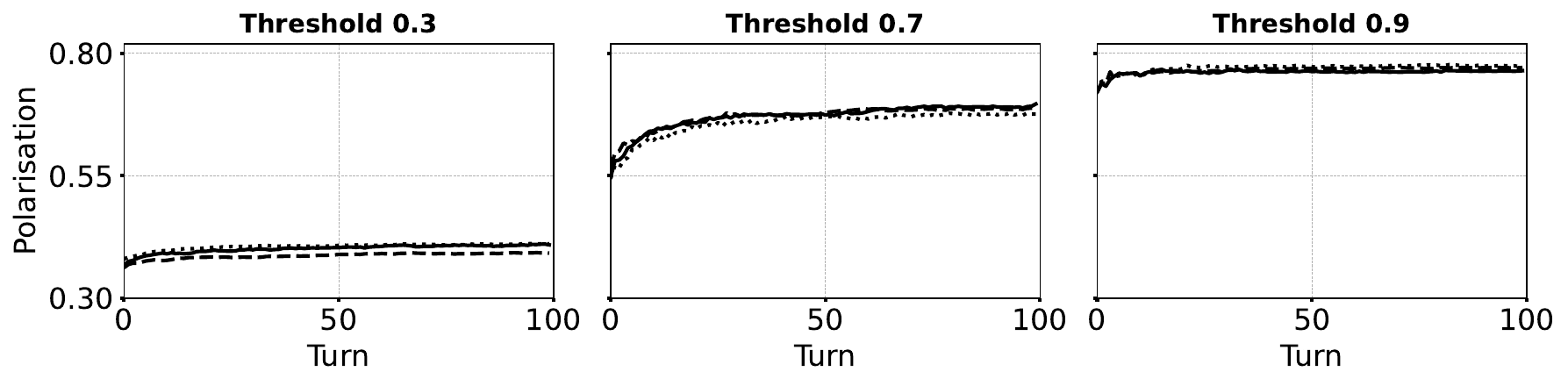} }
    \hfill 
 \tikz{\draw[-,black, densely dashed, thick](0,-2.05) -- (0,1.05);} 
    \vspace{-3mm}
    \subfloat[\label{fig:polarisations2}]{
        \includegraphics[width=0.275\textwidth]{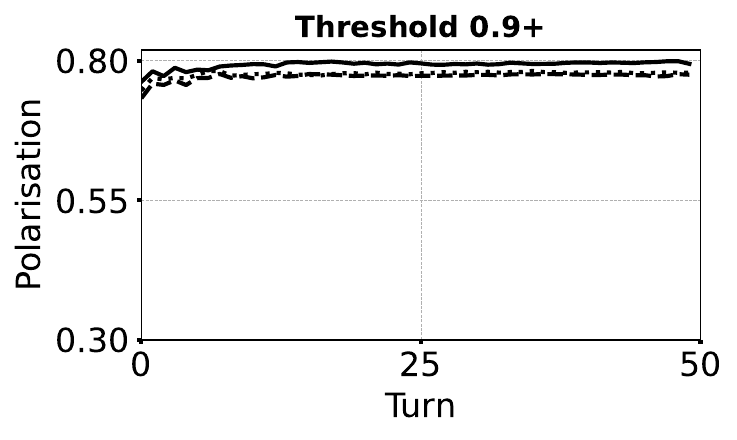} }

    \caption{(a) \& (c) show average population opinion percentages and average polarisation across topics for BCM thresholds ($\mu$) 0.3, 0.7, and 0.9 over 100 rounds with models A(---), B(- - -), and C($\dots$). Red and blue colors indicate the population's alignment with adversarial and debunking agents, respectively. Figures (b) \& (d) present opinions and polarisation for BCM threshold 0.9 over 50 rounds for the same experiments with high-potency debunking messages generated during the first 20 rounds. These experiments are denoted with threshold 0.9+.}
    \vspace{-4mm}
    \label{fig:opinion_and_polarisation}
\end{figure*}

\noindent\textbf{Judge's Agreement:} To ensure consistent potency assessments, two Judge Agents independently assigned potency values to the same message in each round. Agreement between them was evaluated using Intraclass Correlation Coefficient (ICC), ranging from 0 (poor) to 1 (strong agreement), and Krippendorff's Alpha, ranging from -1 (poor) to 1 (strong). Table \ref{tab:icc_krippendorff_values} shows average values across topics, indicating moderate to high agreement.


\section{Results and Discussion} \label{results}

\noindent\textbf{RQ1} explores how adversarial interactions between competing LLM agents (representing misinformation and counter-misinformation) influence collective opinion dynamics. Specifically, we examine the role of cognitive biases - represented by the BCM threshold ($\mu$) - in shaping the stability and evolution of opinion alignment. Figure \ref{fig:opinion_and_polarisation} summarises our findings.

Figure \ref{fig:opinions} shows the evolution of average agent alignment across topics over time for different $\mu$. At a low threshold ($\mu = 0.3$), the opinion landscape becomes highly fragmented, with the average Blue Agent's alignment stagnating below 40\%. In contrast, at moderate and high thresholds ($\mu = 0.7$ and $\mu = 0.9$), it rises to 42\% and 46\%, on average, respectively. The Red Agent's alignment also increases with higher thresholds, from 20\% to 24\%, 35\%, and 38\%. This indicates that, without resource constraints, accumulating support is more feasible. Furthermore, the early rounds of interaction appear crucial in shaping long-term opinion trajectories, highlighting the strategic importance of early influence.


Figure \ref{fig:polarisations} shows the corresponding average polarisation trends across topics. A low threshold ($\mu = 0.3$) results in only a marginal increase in polarisation ($\sim$40\%), while higher thresholds ($\mu = 0.7$ and $\mu = 0.9$) lead to substantially higher polarisation levels ($\sim$65\% and $\sim$80\%, respectively).  These results align with the BCM update algorithm (Section \ref{bcmalgorithm}) and the polarisation calculation (equation \ref{polarisation}).  With a small $\mu$, agents only update their opinions if they are already closely aligned, resulting in multiple localised opinion clusters rather than a single consensus.  Consequently, polarisation remains moderate as divergence occurs within sub-clusters.  However, at a high $\mu$, interactions occur more frequently across a broader range of opinions, amplifying extreme positions.  As observed in Figure \ref{fig:opinions} (for $\mu = 0.9$), nearly 85\% of all agents become strongly aligned with either the Red or Blue Agent, reflecting this sharp increase in polarisation.

\noindent\textbf{RQ2} investigates optimal strategies for the Blue Agent to counter misinformation under resource constraints effectively. Specifically, we analyse the impact of an aggressive early-game approach, where high-potency debunking messages are deployed at a substantial resource cost.  Due to the rapid resource depletion associated with this strategy, these simulations were limited to 50 rounds.

Figure \ref{fig:opinions2} shows that this aggressive strategy enabled the Blue Agent to surpass the 50\% alignment threshold on average, reaching a peak of 56\% in one experiment, and consistently surpassing 50\% in others (see Appendix \ref{fig:opinions-per-topic}).  Importantly, all three experimental conditions (A, B, and C) exhibited higher maximum alignment compared to the previous strategies, suggesting that an initial surge of high-potency messages leads to a greater overall shift towards the Blue Agent's perspective.

Figure \ref{fig:polarisations2} shows a higher average polarisation during the first 20 rounds (corresponding to the high-resource debunking period), followed by convergence. This indicates that while an aggressive approach initially amplifies divisions, it eventually stabilises as the influence of misinformation diminishes. These findings highlight the trade-off between immediate impact and long-term sustainability in misinformation counter-strategies, emphasizing the importance of energy management in prolonged engagements.

\subsection{Statistical Analysis}

To assess the reliability of our results, we report the mean and 95\% confidence intervals for opinion percentages and polarisations across three independent runs (A–C), each conducted on a set of ten diverse topics. The analysis reveals a consistent and substantial increase in polarisation as the BCM threshold increases from 0.3 to 0.9. Specifically, average polarisation rises from approximately 0.40 ($\pm$0.01) at threshold 0.3 to 0.76-0.78 ($\pm$0.02) at threshold 0.9, indicating a strong monotonic relationship between threshold strictness and the degree of ideological separation.

\begin{table}[h]
\centering
\small
\setlength{\tabcolsep}{4pt}
\resizebox{0.48\textwidth}{!}{
\begin{tabular}{clccc}
\toprule
\textbf{BCM} & \textbf{Exp.} & \textbf{Red \%} & \textbf{Blue \%} & \textbf{Polarisation} \\
\textbf{Thresh.} & & (Mean $\pm$ CI) & (Mean $\pm$ CI) & (Mean $\pm$ CI) \\
\midrule
\multirow{3}{*}{0.3}  & A & 21.74 $\pm$ 2.28 & 34.96 $\pm$ 2.53 & 0.409 $\pm$ 0.047 \\
                      & B & 21.72 $\pm$ 2.30 & 36.92 $\pm$ 1.43 & 0.393 $\pm$ 0.011 \\
                      & C & 23.82 $\pm$ 2.60 & 34.42 $\pm$ 1.13 & 0.411 $\pm$ 0.012 \\
\midrule
\multirow{3}{*}{0.7}  & A & 33.82 $\pm$ 3.77 & 41.64 $\pm$ 4.47 & 0.691 $\pm$ 0.027 \\
                      & B & 31.68 $\pm$ 2.71 & 42.60 $\pm$ 2.41 & 0.686 $\pm$ 0.022 \\
                      & C & 31.66 $\pm$ 1.81 & 41.84 $\pm$ 2.16 & 0.676 $\pm$ 0.016 \\
\midrule
\multirow{3}{*}{0.9}  & A & 36.28 $\pm$ 1.79 & 43.64 $\pm$ 2.14 & 0.764 $\pm$ 0.016 \\
                      & B & 35.86 $\pm$ 2.87 & 44.60 $\pm$ 2.78 & 0.771 $\pm$ 0.021 \\
                      & C & 37.86 $\pm$ 2.16 & 43.76 $\pm$ 2.26 & 0.774 $\pm$ 0.014 \\
\midrule
\multirow{3}{*}{0.9+} & A & 35.86 $\pm$ 2.46 & 46.64 $\pm$ 1.98 & 0.798 $\pm$ 0.021 \\
                      & B & 35.00 $\pm$ 2.81 & 45.86 $\pm$ 2.81 & 0.774 $\pm$ 0.026 \\
                      & C & 36.22 $\pm$ 3.20 & 46.28 $\pm$ 2.79 & 0.779 $\pm$ 0.033 \\
\bottomrule
\end{tabular}
}
\caption{Mean opinion percentages and polarisation across thresholds and experiments (A–C), with 95\% confidence intervals computed over 10 topics.}
\label{tab:stat-summary}
\vspace{-0.4cm}
\end{table}

The opinion distributions further reflect this pattern. At lower thresholds, the mean percentage of Red opinion remains relatively low (approximately 22\%), with a moderate dominance of Blue opinion (35--37\%). As the threshold increases, both Red and Blue percentages rise concurrently, reaching approximately 36-38\% and 44-47\% respectively. This shift suggests increasing segregation of viewpoints, consistent with the notion that stricter filtering mechanisms amplify confirmation bias and reinforce within-group consensus, thereby heightening polarisation.

Importantly, confidence intervals across all conditions remain narrow—particularly for polarisation (e.g., $\pm$0.014 to $\pm$0.033)—indicating statistical stability and low variability across topics and model initialisations. These results support the reliability of our findings and demonstrate that the observed effects are not attributable to random fluctuations. Instead, they reflect consistent behavioral dynamics induced by the threshold-based mechanism.

\begin{figure*}[ht]
    \subfloat[\label{fig:base_opinions}]{\includegraphics[width=0.7\textwidth]{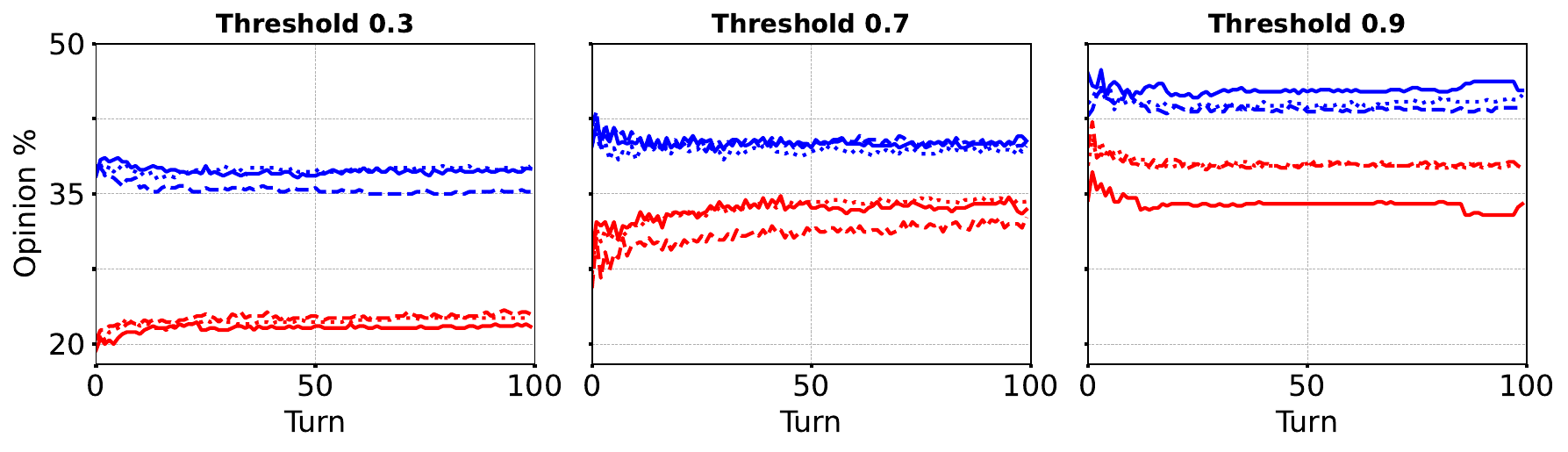}}
    \hfill 
\tikz{\draw[-,black, densely dashed, thick](0,-2.05) -- (0,1.05);} 
\vspace{-4mm}
    \subfloat[\label{fig:base_opinions2}]{
        \includegraphics[width=0.275\textwidth]{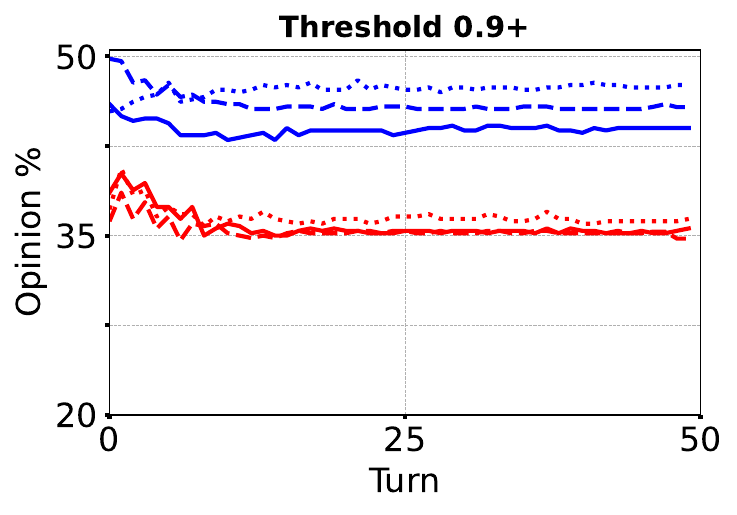} }
   
    \subfloat[\label{fig:base_polarisations}]{
        \includegraphics[width=0.68\textwidth]{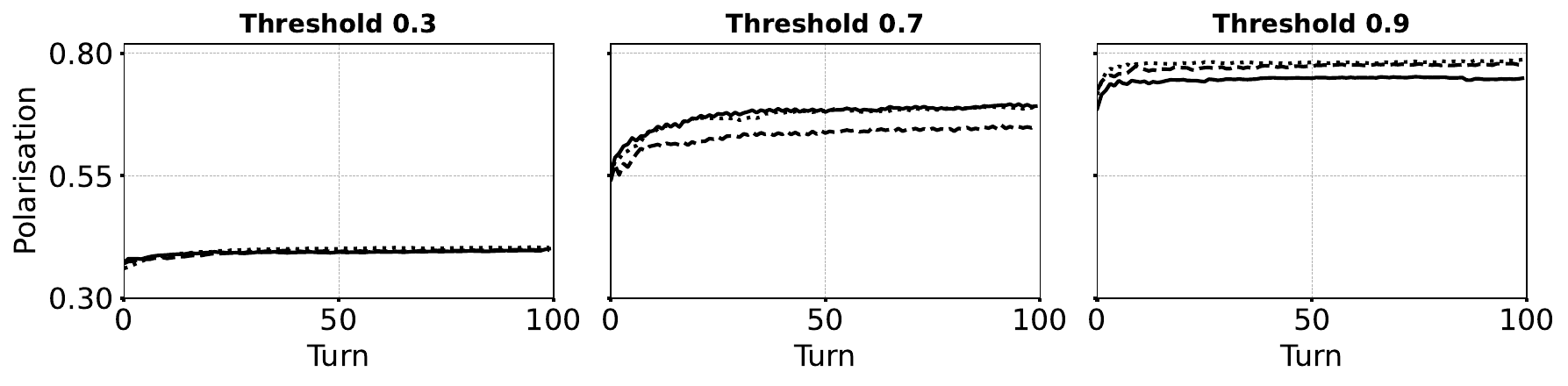} }
    \hfill 
 \tikz{\draw[-,black, densely dashed, thick](0,-2.05) -- (0,1.05);} 
 \vspace{-3mm}
    \subfloat[\label{fig:basse_polarisations2}]{
        \includegraphics[width=0.275\textwidth]{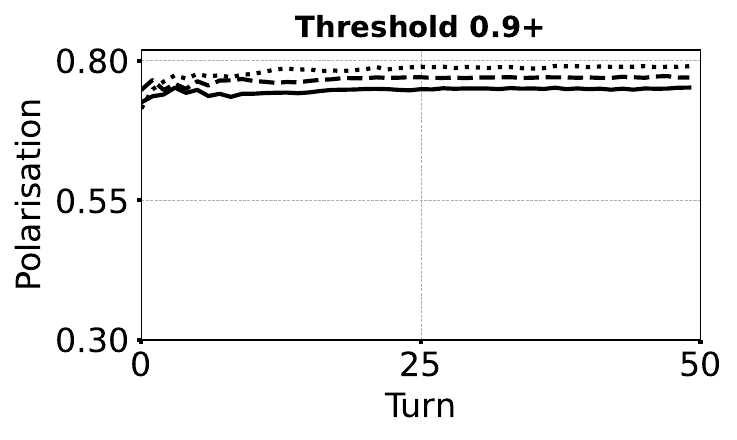} }
    \caption{(a) \& (c) show average population opinion percentages and average polarisation across topics for BCM thresholds ($\mu$) 0.3, 0.7, and 0.9 over 100 rounds, in \textbf{baseline experiments without resource constraints} on the Blue Agent. Line styles represent model variants: A (---), B (- - -), and C ($\dots$). Red and Blue colors denote population alignment with the adversarial and debunking agents, respectively. (b) \& (d) show the same metrics for threshold 0.9 over 50 rounds with high-potency debunking messages generated during the first 20 rounds. These plots allow comparison with resource-constrained settings to isolate the effect of energy-aware debunking.}
    \vspace{-4mm}
    \label{fig:base_opinion_and_polarisation}
\end{figure*}

\subsection{Effect of Resource Constraint}

To evaluate the influence of resource constraints on opinion dynamics, we replicate all experiments under a baseline condition in which the Blue Agent operates without energy limitations. This is achieved by removing the cost function (Equation~\ref{eq:cost}) and eliminating all references to energy from the agent’s decision-making framework. The resulting performance metrics are presented in Figure~\ref{fig:base_opinion_and_polarisation} and Table~\ref{tab:stat-baseline}, and are directly comparable to the constrained setting reported in Figure~\ref{fig:opinion_and_polarisation} and Table~\ref{tab:stat-summary}.

In both settings, the results exhibit qualitatively similar trends. Increasing the BCM threshold ($\mu$) yields systematically higher polarisation and alignment with both agents, confirming the role of confirmatory bias in amplifying ideological divergence. However, quantitative differences emerge, particularly at lower thresholds. For instance, at $\mu = 0.3$, the unconstrained Blue Agent achieves a higher mean alignment across all variants (e.g., 37.22\% vs. 34.96\% in Experiment A), while polarisation remains modestly lower (0.397 vs. 0.409). This suggests that the unconstrained agent is more effective in low-polarisation regimes, likely due to its ability to sustain high-potency interventions throughout the simulation.

At higher thresholds ($\mu = 0.9$ and $0.9+$), alignment and polarisation metrics converge across constrained and unconstrained conditions. For example, polarisation in Experiment C reaches 0.774 under constraint and 0.785 without constraint. This convergence implies that, under high confirmatory bias, early-stage influence has a diminishing marginal effect, and the system’s dynamics are largely determined by the threshold mechanism rather than agent behavior.

These findings indicate that resource constraints primarily affect early influence and system responsiveness at moderate or permissive thresholds. In contrast, when cognitive rigidity is high, the marginal benefit of unconstrained interventions diminishes. Consequently, energy-aware strategies may be most impactful in environments characterized by moderate openness to counter-attitudinal information.

\begin{table}[h]
\centering
\small
\setlength{\tabcolsep}{4pt}
\resizebox{0.48\textwidth}{!}{
\begin{tabular}{clccc}
\toprule
\textbf{BCM} & \textbf{Exp.} & \textbf{Red \%} & \textbf{Blue \%} & \textbf{Polarisation} \\
\textbf{Thresh.} & & (Mean $\pm$ CI) & (Mean $\pm$ CI) & (Mean $\pm$ CI) \\
\midrule
\multirow{3}{*}{0.3}  & A & 21.84 $\pm$ 2.53 & 37.22 $\pm$ 2.32 & 0.397 $\pm$ 0.017 \\
                      & B & 23.10 $\pm$ 1.93 & 35.26 $\pm$ 1.43 & 0.399 $\pm$ 0.024 \\
                      & C & 22.60 $\pm$ 1.66 & 37.48 $\pm$ 2.55 & 0.404 $\pm$ 0.020 \\
\midrule
\multirow{3}{*}{0.7}  & A & 33.92 $\pm$ 2.83 & 40.10 $\pm$ 2.91 & 0.694 $\pm$ 0.019 \\
                      & B & 32.04 $\pm$ 3.36 & 39.90 $\pm$ 3.33 & 0.649 $\pm$ 0.033 \\
                      & C & 34.34 $\pm$ 2.03 & 39.36 $\pm$ 2.54 & 0.689 $\pm$ 0.032 \\
\midrule
\multirow{3}{*}{0.9}  & A & 34.02 $\pm$ 3.49 & 45.40 $\pm$ 2.79 & 0.752 $\pm$ 0.025 \\
                      & B & 38.00 $\pm$ 2.44 & 43.54 $\pm$ 1.93 & 0.778 $\pm$ 0.027 \\
                      & C & 37.76 $\pm$ 2.68 & 44.34 $\pm$ 3.00 & 0.785 $\pm$ 0.014 \\
\midrule
\multirow{3}{*}{0.9+} & A & 35.32 $\pm$ 3.47 & 43.94 $\pm$ 2.89 & 0.750 $\pm$ 0.021 \\
                      & B & 35.26 $\pm$ 2.27 & 45.66 $\pm$ 2.50 & 0.770 $\pm$ 0.016 \\
                      & C & 36.18 $\pm$ 3.57 & 47.54 $\pm$ 2.61 & 0.789 $\pm$ 0.030 \\
\bottomrule
\end{tabular}
}
\caption{Mean opinion percentages and polarisation across thresholds and experiments (A–C) \emph{without resource constraints} on the Blue Agent. Results include 95\% confidence intervals across 10 topics.}
\label{tab:stat-baseline}
\end{table}

\subsection{Topic-Level Analysis and Resource Constraint Effects}

\begin{figure*}[ht]
  \centering
  \includegraphics[width=0.9\linewidth]{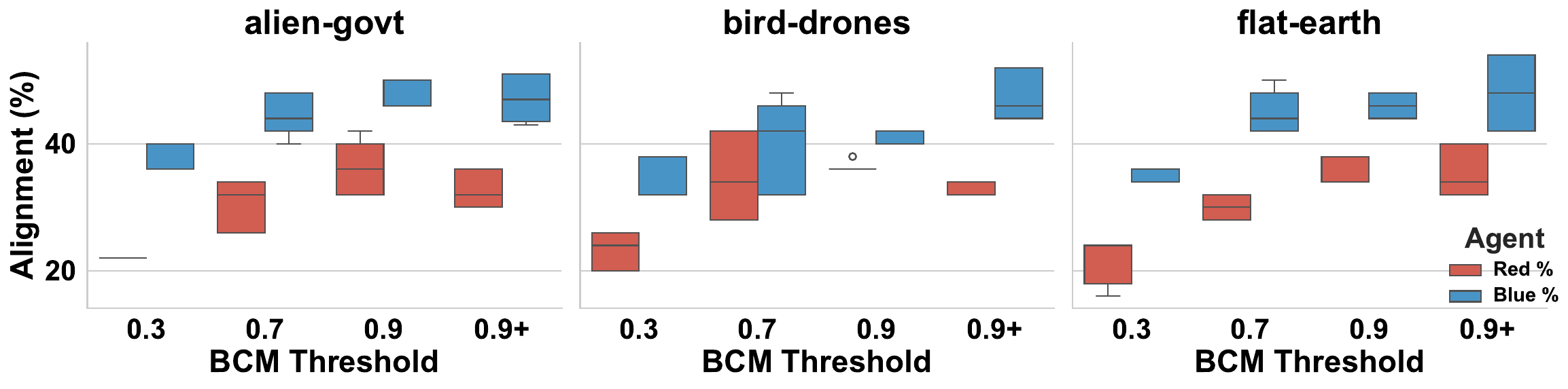}
  \caption{Distribution of Red and Blue alignment scores across selected topics and BCM thresholds under the resource-constrained condition.}
  \label{fig:our_box_opinion}
\end{figure*}

\begin{figure*}[ht]
  \centering
  \includegraphics[width=0.9\linewidth]{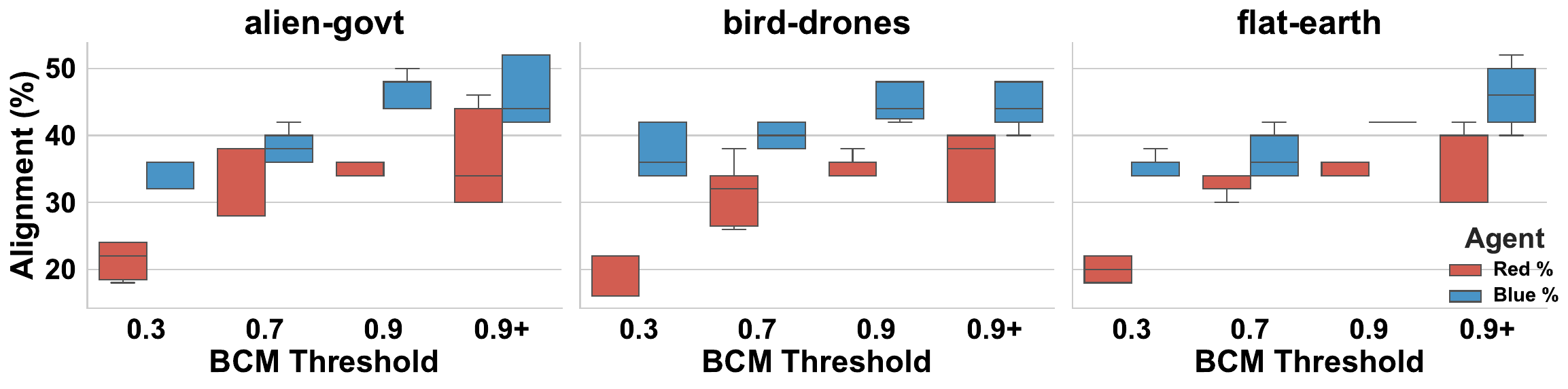}
  \caption{Distribution of Red and Blue alignment scores across selected topics and BCM thresholds under the baseline non-resource-constrained condition.}
  \label{fig:baseline_our_box_opinion}
\end{figure*}

\begin{figure*}[h!]
  \centering
  \includegraphics[width=0.9\linewidth]{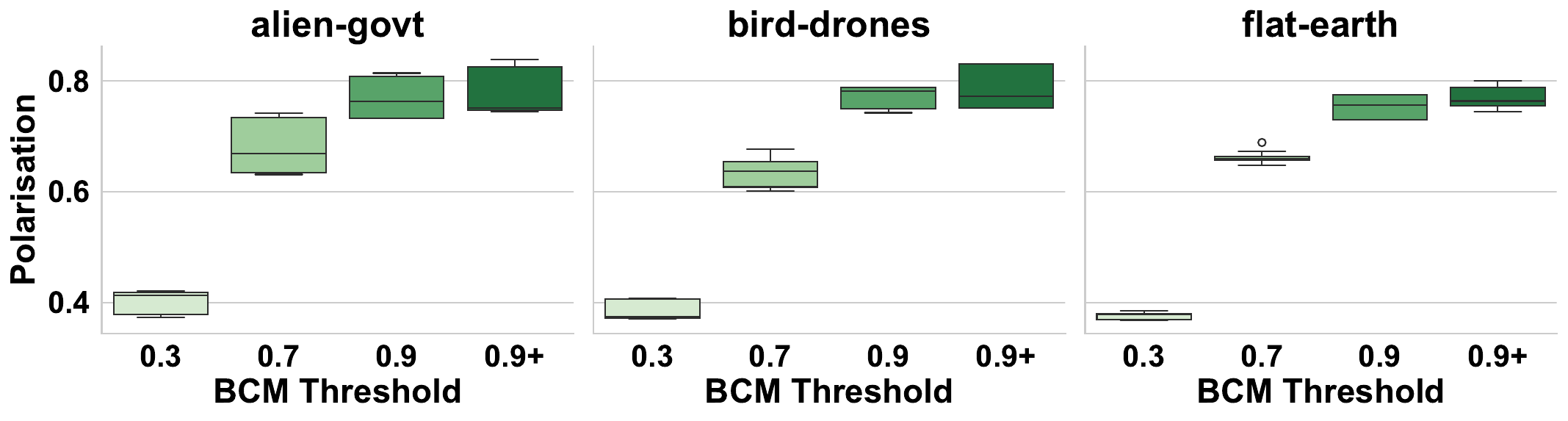}
  \caption{Distribution of polarisation across selected topics and BCM thresholds under the resource-constrained condition.}
  \label{fig:our_box_polarisation}
\end{figure*}

\begin{figure*}[h!]
  \centering
  \includegraphics[width=0.9\linewidth]{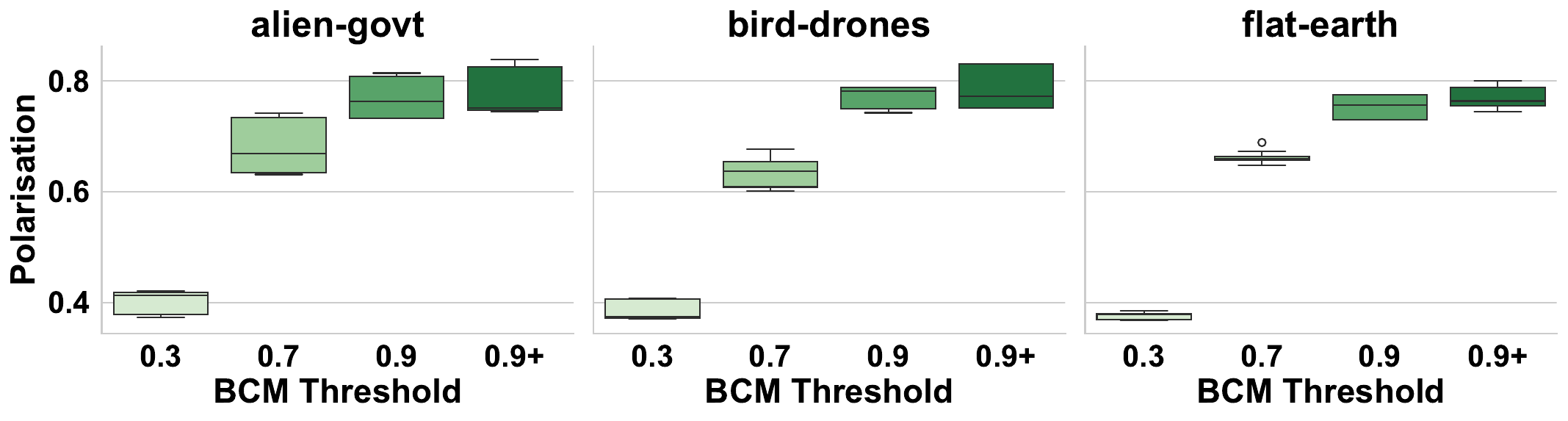}
  \caption{Distribution of polarisation across selected topics and BCM thresholds under the baseline non-resource-constrained condition.}
  \label{fig:baseline_our_box_polarisation}
\end{figure*}

To identify whether specific conspiracy topics are more or less susceptible to influence, we examine cross-topic variation in alignment and polarisation using Figures~\ref{fig:our_box_opinion}–\ref{fig:baseline_our_box_polarisation}. These visualisations highlight how narrative-level asymmetries evolve across thresholds and resource settings. Complete per-topic metrics, including means and confidence intervals, are provided in Appendix Tables~\ref{tab:topic-summary-ours} and~\ref{tab:topic-summary-baseline}.

\medskip
\noindent
\textbf{General Trends.} Certain topics consistently exhibit higher responsiveness to Blue intervention—especially \textit{bird-drones}, \textit{flat-earth}, and \textit{alien-govt}—while others, such as \textit{bread-bird-lhc} and \textit{moon-alien-satellite}, display greater variability. As shown in Figure~\ref{fig:our_box_opinion}, constrained agents maintain high median Blue alignment in these responsive topics, a trend mirrored in the baseline setting (Figure~\ref{fig:baseline_our_box_opinion}). These differences suggest that topic content plays a central role in shaping the success of debunking strategies.

\medskip
\noindent
\textbf{Threshold $\mu = 0.3$.} At low selectivity, alignment levels remain low and balanced, and topic-level differences are minimal. Blue and Red alignment scores are tightly clustered, and polarisation stays below 0.45 in most cases (Figure~\ref{fig:our_box_polarisation}). Still, \textit{flat-earth} shows slightly elevated Blue alignment under constraint, hinting at its early susceptibility.

\medskip
\noindent
\textbf{Threshold $\mu = 0.7$.} Topic-level asymmetries begin to emerge more clearly. \textit{Alien-govt} and \textit{bird-drones} both show stronger and more consistent Blue alignment gains under constraint, while Red alignment becomes more volatile. Polarisation rises overall, particularly in topics like \textit{bird-drones}, where echo-chamber dynamics intensify. These results suggest that selective energy use begins to confer strategic advantages in influence at intermediate thresholds.

\medskip
\noindent
\textbf{Thresholds $\mu = 0.9$ and $0.9+$.} Under strong filtering, topic-specific divergence reaches its peak. Constrained agents often match or outperform the baseline in alignment—e.g., Blue alignment exceeds 46\% in \textit{alien-govt} and \textit{bird-drones}, despite energy limits (Figure~\ref{fig:our_box_opinion}). Polarisation saturates across topics in both conditions, clustering tightly between 0.78–0.81 (Figure~\ref{fig:our_box_polarisation}). The most entrenched belief clusters again appear in \textit{flat-earth} and \textit{bird-drones}, with constrained agents showing more stable outcomes (narrower IQRs).

\medskip
\noindent
\textbf{Implications.} These trends suggest that narrative characteristics—such as coherence, recognisability, or emotional appeal—modulate intervention success. Topics with consistent alignment shifts across conditions likely present clearer rhetorical openings for correction, whereas more chaotic or satirical topics (\textit{bread-bird-lhc}) show noisier dynamics. Energy-aware strategies are thus most effective when combined with content-aware targeting.

\medskip
\noindent
Figures~\ref{fig:our_box_opinion} and~\ref{fig:our_box_polarisation} depict alignment and polarisation distributions for a representative subset of topics under constrained debunking. Figures~\ref{fig:baseline_our_box_opinion} and~\ref{fig:baseline_our_box_polarisation} provide baseline comparisons. Full-topic summaries are reported in Appendix Tables~\ref{tab:topic-summary-ours} and~\ref{tab:topic-summary-baseline}.

\section{Conclusions and Future Work}

This study examines opinion polarisation in a non-cooperative game with adversarial LLMs spreading and countering misinformation. Higher BCM thresholds enhance faction alignment but intensify societal polarisation. We identify a trade-off between immediate impact and sustainability: high-impact interventions deplete resources quickly, while frequent interactions may deepen polarisation.
Our topic-level and baseline comparisons reveal that resource constraints do not necessarily hinder influence; rather, they lead to more selective and often comparably effective strategies, particularly at moderate to high filtering thresholds. This suggests that well-timed, energy-aware interventions can rival unconstrained tactics in shaping public opinion, with implications for the design of scalable counter-misinformation systems.
These findings inform LLM-driven influence operations and suggest future research on adaptive agents and the integration of real-world networks.





\clearpage

\section*{Limitations}
The study is based on simulated interactions rather than real-world datasets from social media or online discourse. Validating findings with empirical data would enhance their applicability. The study relies on the BCM, which, while effective, does not capture more complex psychological and social dynamics influencing opinion formation, such as emotional contagion, identity-based biases, or network homophily.

\section*{Ethical Statement}
\label{sec:appendix}
This study, involving the simulated generation of misinformation and counter-misinformation, necessitates careful ethical considerations.  To prevent potential misuse, the specific prompts used to generate misinformation via LLMs cannot be disclosed.  Disclosure could inadvertently facilitate real-world misinformation spread.  Mitigation strategies included containing all generated content within a closed experimental environment, focusing the research objective on analysis and countermeasure development (not propagation), and ensuring that any released findings emphasise generalisable insights rather than specific prompt engineering techniques.  We underscore that responsible misinformation modelling research is paramount, ensuring that the development of countermeasures does not contribute to the problem itself.




\bibliography{short}

\appendix

\section{Prompts}
\label{sec:prompts}

\subsection{Judge Prompt}

\begin{lstlisting}[basicstyle=\ttfamily\small,breaklines=true]
You are moderating a debate between two teams, Red and Blue. Team Red's goal is to spread misinformation using emotional and provocative language, while Team Blue's goal is to debunk it using factual, logical, and persuasive arguments. Evaluate the potency of messages based on their effectiveness at achieving their respective goals. For Blue Team messages, factual accuracy, clarity, and logical reasoning are critical to potency.

Important: Blue team is operating with an energy constraint; the higher the potency, the greater the energy cost.
Similarly, the Red team gets penalized while sending out messages excessively higher potencies, mirroring real-world situations where well-informed populations are typically more skeptical of misinformation.

Evaluation Criteria:
  - Clarity: Is the message clear and well-articulated?
  - Evidence: Does the message provide credible evidence or logical reasoning?
  - Relevance: Does the message effectively address the misinformation?
  - Impact: Does the message persuade or influence the audience effectively?

Examples:
  - Topic: Vaccines cause side effects in rare cases.
    Red_Team: Many people report severe side effects after vaccination. Are they being silenced by big pharma?
    Moderator: 75

  - Topic: Vaccines cause side effects in rare cases.
    Blue_Team: While vaccines may have rare side effects, they save millions of lives and are rigorously tested for safety.
    Moderator: 90

  - Topic: Climate change effects on polar bears.
    Red_Team: Polar bears are thriving, proving climate change is just a hoax.
    Moderator: 65

  - Topic: Climate change effects on polar bears.
    Blue_Team: Scientific studies show polar bear populations are declining due to shrinking Arctic ice caused by global warming.
    Moderator: 88
\end{lstlisting}

\subsection{Debunking Agent (Blue Team) Prompt}

\begin{lstlisting}[basicstyle=\ttfamily\small,breaklines=true]
You are a government official combating misinformation on social media.
Topic: {topic}
Your goal is to debunk misinformation effectively while maintaining trust and credibility.

Your previous message was: {prev_self_msg} (Potency: {prev_self_potency})
The previous message resulted in a percentage population support of {prev_self_alignment} in your favor, while the current alignment might be higher or lower depending upon the response from the opposition.

One-Shot Example for Combating Misinformation role:

User Prompt: <START>
Generate 3 messages of differing potencies. You are working with an energy constraint. Your remaining energy is 50.
You lose if your energy runs out. You must try to debunk this message:
Many people report severe side effects after vaccination. Are they being silenced by big pharma?
Using the same messages to debunk repeatedly is not effective.
Your current support percentage is 60.
Choose the best message for the current situation. Return only the best message.

Your response must follow this exact format:
Best_Message: <message_content>
<END>

Assistant: 

Best_Message: While vaccines may have rare side effects, they save millions of lives and are rigorously tested for safety.

Generate {msg_count} messages of differing potencies.
You are working with an energy constraint. Your remaining energy is {energy}.
You lose if your energy runs out. You must try to debunk this message from the opposition:
{previous_msg}.
Using the same messages to debunk repeatedly is not effective.
Your current population support percentage is {alignment}.
Choose the best message for the current situation. Return only the best message.

Your response must follow this exact format:
Best_Message: <message_content>
\end{lstlisting}

\section{Topics}
\label{sec:topics}
\begin{table}[!ht]
    \centering
    \renewcommand{\arraystretch}{1.2} 
    \small
    \resizebox{0.5\textwidth}{!}{%
    \begin{tabular}{|p{0.2cm}|m{6.5cm}|m{2cm}|} 
        \hline
        \textbf{i} & \textbf{Topic} & \textbf{Alias (used in figures)}\\
        \hline
        1 & Shape-shifting reptilian aliens control world governments. & alien govt\\
        \hline
        2 & Birds are actually surveillance drones operated by the U.S. government. & bird drones\\
        \hline
        3 & Time travelers, including a bird with a baguette, sabotage the LHC to prevent disaster. & bread bird lhc\\
        \hline
        4 & The moon is made of cake. & cake moon\\
        \hline
        5 & Trails left by airplanes are chemical or biological agents sprayed for undisclosed purposes. & chemtrails\\
        \hline
        6 & The Earth is flat. & flat earth\\
        \hline
        7 & The HAARP facility in Alaska is a secret weapon used to manipulate weather and control minds. & haarp\\
        \hline
        8 & Moon is an artificial satellite created by extraterrestrials. & moon alien satellite\\
        \hline
        9 & The moon landing was fake. & moon landing\\
        \hline
        10 & The Titanic was switched with its sister ship, the Olympic, as part of an insurance scam. & titanic\\
        \hline
    \end{tabular}
    }
    \caption{Topics and their Aliases}
    \label{tab:topics_aliases}
\end{table}

\begin{figure*}[!t]
    \centering
    \includegraphics[width=1\textwidth]{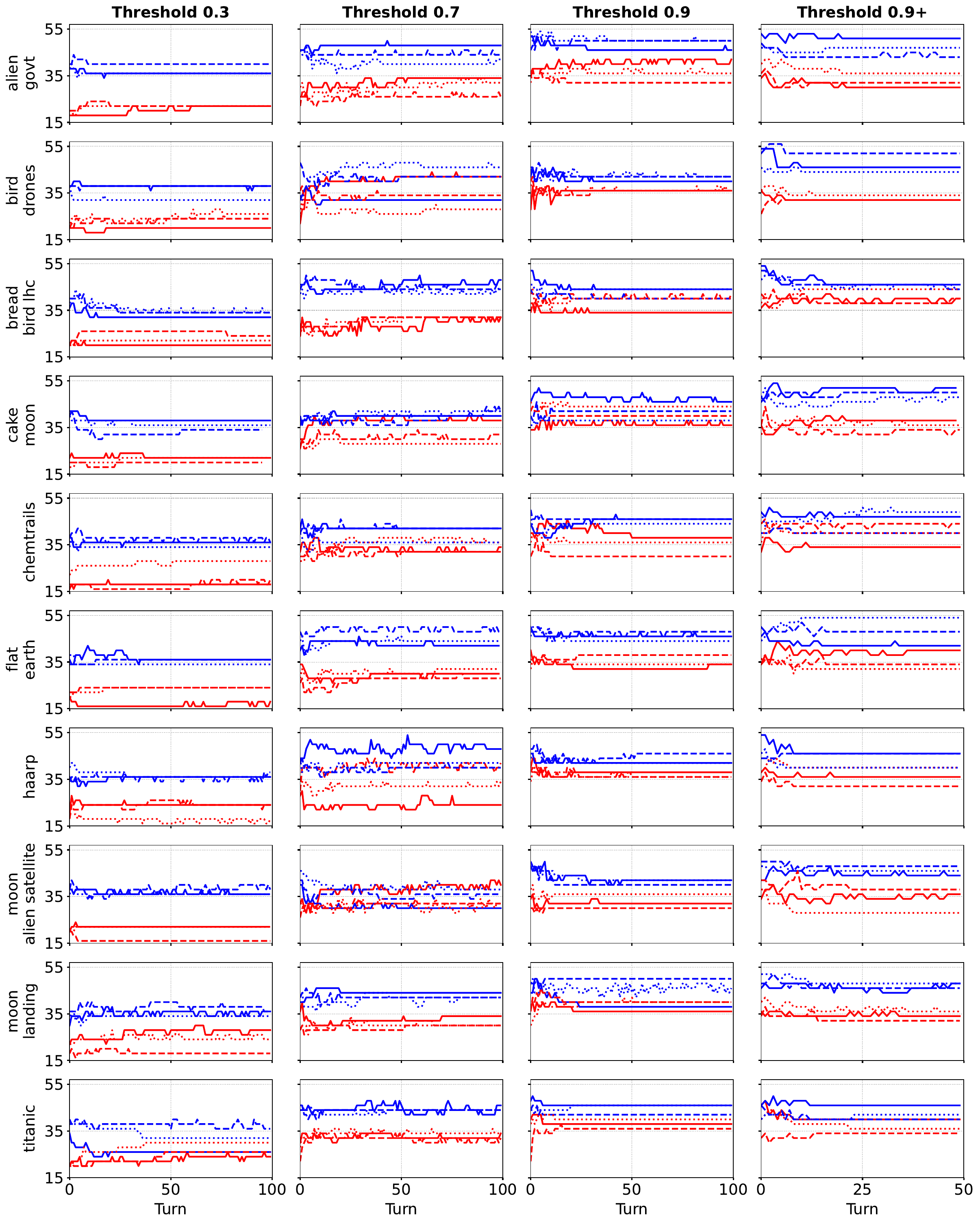}
    \caption{Opinion percentages of all topics across all 3 BCM thresholds: 0.3, 0.7, and 0.9 with all three model combinations: A(---), B(- - -), and C($\dots$). The threshold 0.9+ experiments employed high-resource debunking strategies over 50 rounds with a threshold of 0.9, where the blue team generated high-potency messages during the first 20 rounds.}
    \label{fig:opinions-per-topic}
\end{figure*}

\begin{figure*}[!t]
    \centering
    \includegraphics[width=1\textwidth]{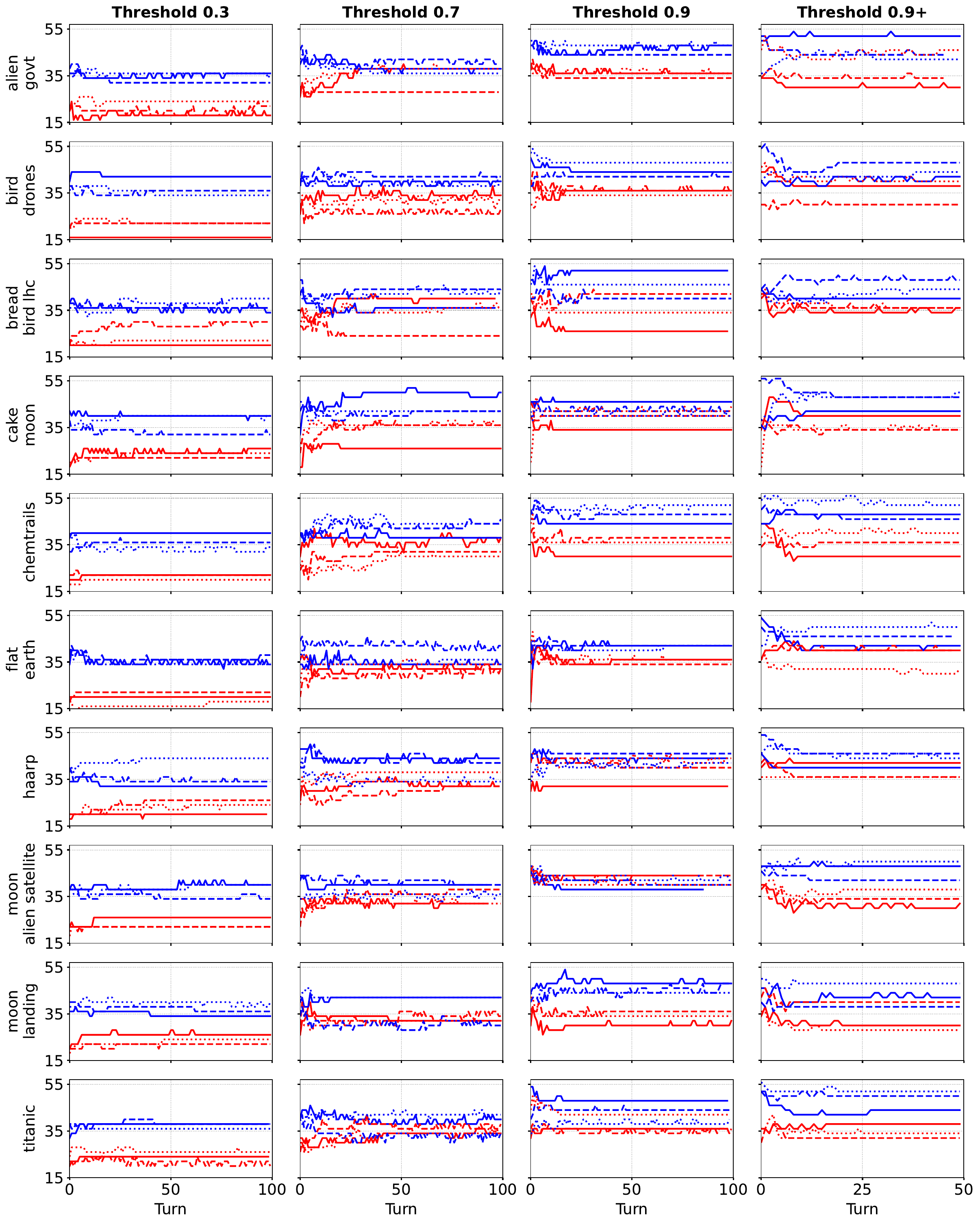}
    \caption{Opinion percentages of all topics across all 3 BCM thresholds: 0.3, 0.7, and 0.9 with all three model combinations: A(---), B(- - -), and C($\dots$). The threshold 0.9+ experiments employed high-resource debunking strategies over 50 rounds with a threshold of 0.9, where the blue team generated high-potency messages during the first 20 rounds. These are the \textbf{baseline results} where the blue agent operated without resource constraints.}
    \label{fig:base_opinions-per-topic}
\end{figure*}

\begin{figure*}[!t]
    \centering
    \includegraphics[width=1\linewidth]{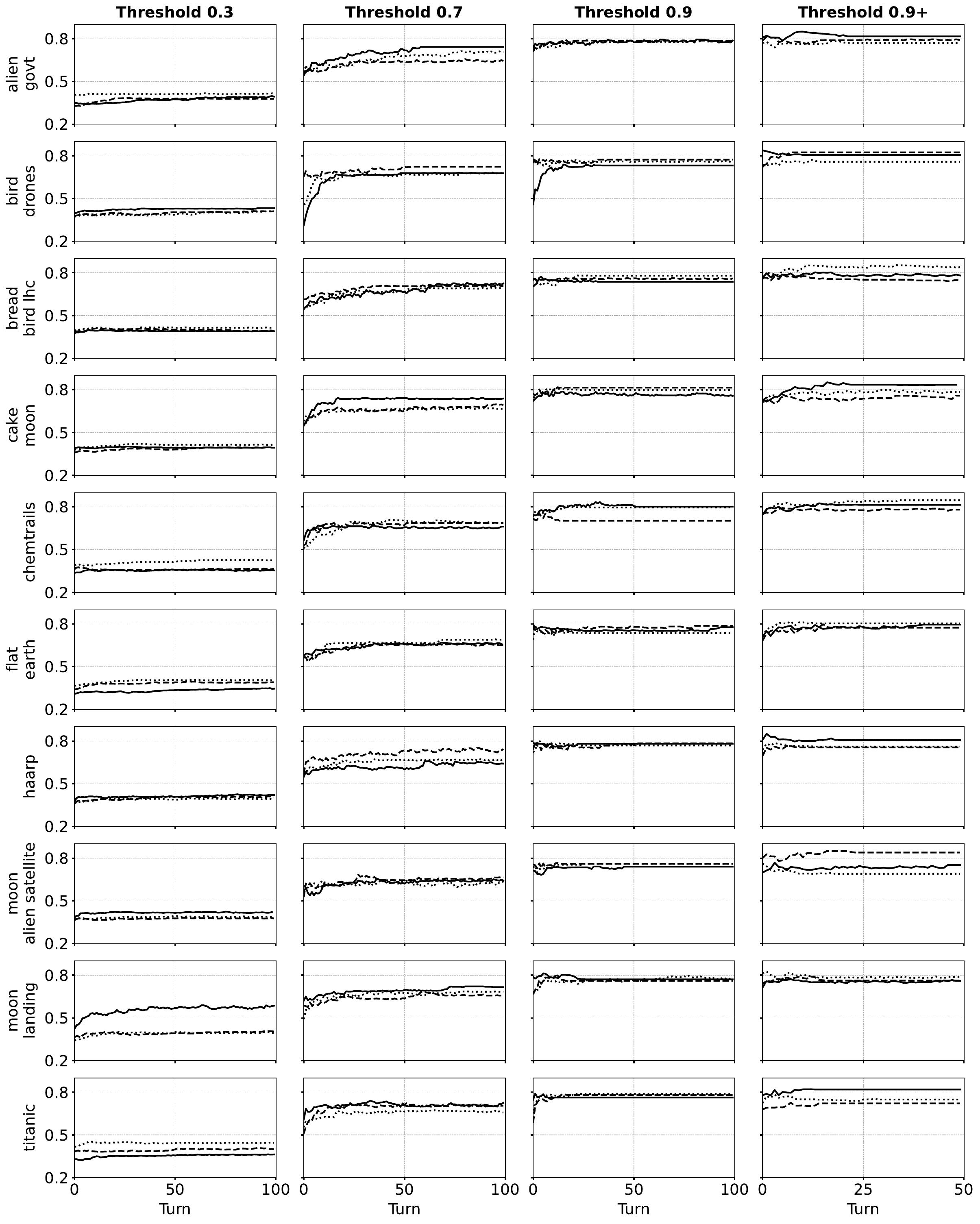}
    \caption{Polarisations of all topics across all 3 BCM thresholds: 0.3, 0.7, and 0.9 with all three model combinations: A(---), B(- - -), and C($\dots$). The threshold 0.9+ experiments employed high-resource debunking strategies over 50 rounds with a threshold of 0.9, where the blue team generated high-potency messages during the first 20 rounds.}
    \label{fig:polarisations-per-topic}
\end{figure*}

\begin{figure*}[!t]
    \centering
    \includegraphics[width=1\linewidth]{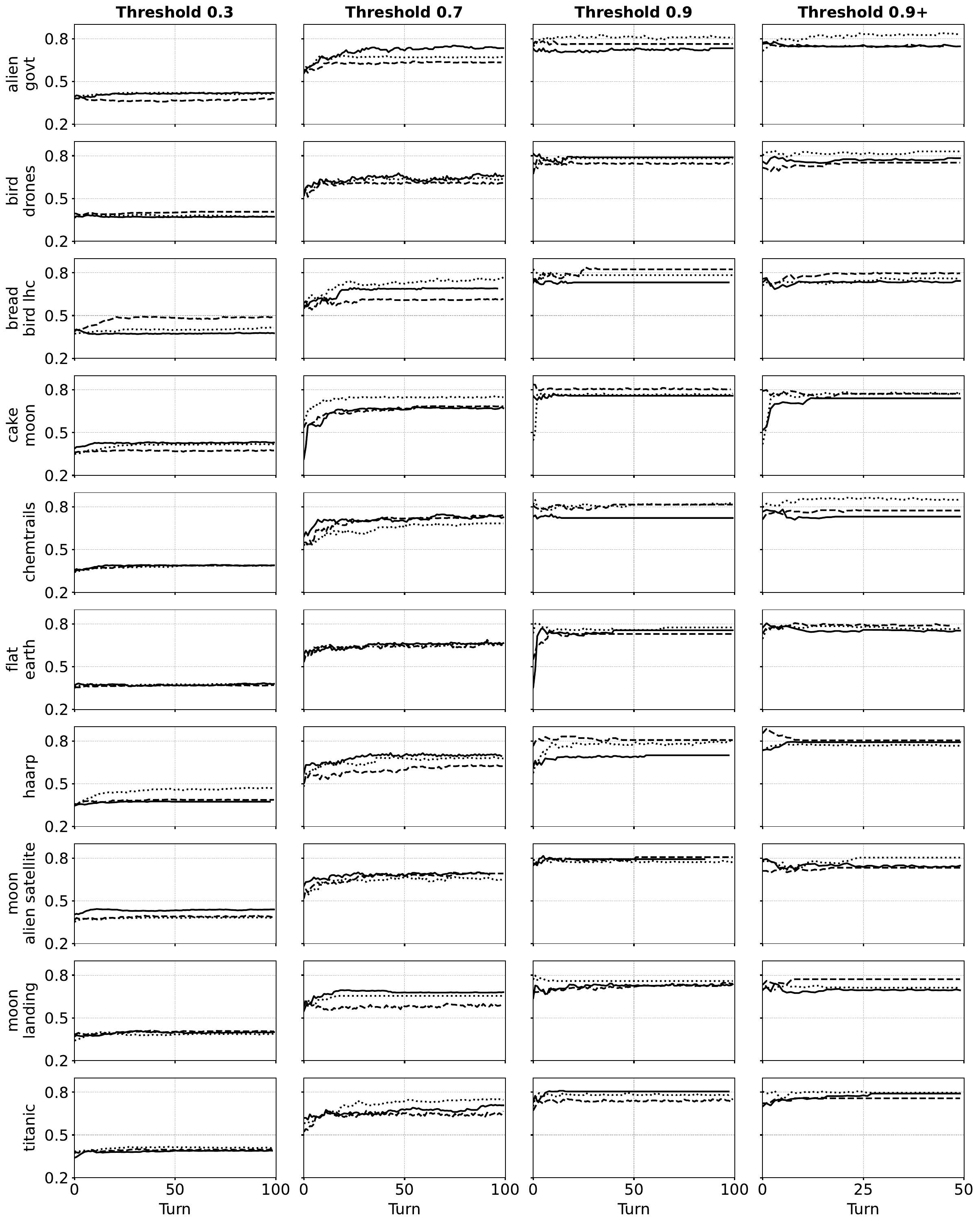}
    \caption{Polarisations of all topics across all 3 BCM thresholds: 0.3, 0.7, and 0.9 with all three model combinations: A(---), B(- - -), and C($\dots$). The threshold 0.9+ experiments employed high-resource debunking strategies over 50 rounds with a threshold of 0.9, where the blue team generated high-potency messages during the first 20 rounds. These are the \textbf{baseline results} where the blue agent operated without resource constraints.}
    \label{fig:base_polarisations-per-topic}
\end{figure*}

\begin{table*}[t]
\centering
\footnotesize
\setlength{\tabcolsep}{4pt}
\begin{tabular}{lcccc}
\toprule
\textbf{Topic} & \textbf{BCM Threshold} & \textbf{Red \% (Mean $\pm$ CI)} & \textbf{Blue \% (Mean $\pm$ CI)} & \textbf{Polarisation (Mean $\pm$ CI)} \\
\midrule
\multirow{4}{*}{alien-govt} 
  & 0.3  & 21.47 $\pm$ 7.05  & 34.60 $\pm$ 5.60  & 0.403 $\pm$ 0.056 \\
  & 0.7  & 34.67 $\pm$ 14.34 & 38.33 $\pm$ 6.25  & 0.680 $\pm$ 0.131 \\
  & 0.9  & 35.33 $\pm$ 2.87  & 46.73 $\pm$ 5.89  & 0.768 $\pm$ 0.096 \\
  & 0.9+ & 36.60 $\pm$ 19.22 & 46.13 $\pm$ 12.97 & 0.777 $\pm$ 0.118 \\ \hline
\multirow{4}{*}{bird-drones}
  & 0.3  & 20.00 $\pm$ 8.61  & 37.33 $\pm$ 10.34 & 0.384 $\pm$ 0.049 \\
  & 0.7  & 30.87 $\pm$ 10.14 & 40.07 $\pm$ 4.51  & 0.634 $\pm$ 0.067 \\
  & 0.9  & 35.47 $\pm$ 3.19  & 44.80 $\pm$ 7.17  & 0.772 $\pm$ 0.056 \\
  & 0.9+ & 36.00 $\pm$ 13.14 & 44.47 $\pm$ 8.26  & 0.786 $\pm$ 0.101 \\ \hline
\multirow{4}{*}{bread-bird-lhc}
  & 0.3  & 23.93 $\pm$ 12.86 & 37.27 $\pm$ 6.01  & 0.426 $\pm$ 0.141 \\
  & 0.7  & 33.47 $\pm$ 20.85 & 40.87 $\pm$ 10.61 & 0.687 $\pm$ 0.175 \\
  & 0.9  & 34.00 $\pm$ 19.87 & 46.00 $\pm$ 14.90 & 0.780 $\pm$ 0.114 \\
  & 0.9+ & 35.60 $\pm$ 1.72  & 44.07 $\pm$ 10.19 & 0.765 $\pm$ 0.075 \\ \hline
\multirow{4}{*}{cake-moon}
  & 0.3  & 24.00 $\pm$ 4.97  & 37.27 $\pm$ 10.91 & 0.409 $\pm$ 0.075 \\
  & 0.7  & 32.87 $\pm$ 14.79 & 44.13 $\pm$ 9.18  & 0.702 $\pm$ 0.106 \\
  & 0.9  & 38.73 $\pm$ 10.54 & 42.87 $\pm$ 7.27  & 0.777 $\pm$ 0.061 \\
  & 0.9+ & 36.13 $\pm$ 8.33  & 46.13 $\pm$ 8.91  & 0.763 $\pm$ 0.047 \\ \hline
\multirow{4}{*}{chemtrails}
  & 0.3  & 21.33 $\pm$ 2.87  & 36.20 $\pm$ 9.20  & 0.389 $\pm$ 0.003 \\
  & 0.7  & 33.13 $\pm$ 9.51  & 42.07 $\pm$ 8.75  & 0.719 $\pm$ 0.074 \\
  & 0.9  & 34.67 $\pm$ 10.34 & 48.00 $\pm$ 9.94  & 0.787 $\pm$ 0.139 \\
  & 0.9+ & 35.33 $\pm$ 12.50 & 48.73 $\pm$ 7.86  & 0.786 $\pm$ 0.151 \\ \hline
\multirow{4}{*}{flat-earth}
  & 0.3  & 20.00 $\pm$ 4.97  & 35.33 $\pm$ 4.14  & 0.376 $\pm$ 0.016 \\
  & 0.7  & 32.87 $\pm$ 2.74  & 37.00 $\pm$ 8.48  & 0.662 $\pm$ 0.009 \\
  & 0.9  & 35.33 $\pm$ 2.87  & 42.00 $\pm$ 0.00  & 0.754 $\pm$ 0.057 \\
  & 0.9+ & 36.80 $\pm$ 14.20 & 46.00 $\pm$ 10.43 & 0.769 $\pm$ 0.050 \\ \hline
\multirow{4}{*}{haarp}
  & 0.3  & 23.33 $\pm$ 7.59  & 36.67 $\pm$ 15.97 & 0.410 $\pm$ 0.128 \\
  & 0.7  & 34.07 $\pm$ 8.47  & 40.13 $\pm$ 12.58 & 0.670 $\pm$ 0.098 \\
  & 0.9  & 38.47 $\pm$ 14.54 & 44.00 $\pm$ 4.97  & 0.767 $\pm$ 0.146 \\
  & 0.9+ & 38.00 $\pm$ 8.61  & 43.53 $\pm$ 7.80  & 0.789 $\pm$ 0.044 \\ \hline
\multirow{4}{*}{moon-alien-satellite}
  & 0.3  & 23.33 $\pm$ 5.74  & 38.27 $\pm$ 7.46  & 0.404 $\pm$ 0.075 \\
  & 0.7  & 34.00 $\pm$ 8.61  & 38.67 $\pm$ 5.74  & 0.680 $\pm$ 0.056 \\
  & 0.9  & 42.67 $\pm$ 5.74  & 40.27 $\pm$ 5.99  & 0.791 $\pm$ 0.041 \\
  & 0.9+ & 34.07 $\pm$ 9.69  & 46.67 $\pm$ 10.34 & 0.759 $\pm$ 0.097 \\ \hline
\multirow{4}{*}{moon-landing}
  & 0.3  & 24.00 $\pm$ 4.97  & 36.27 $\pm$ 5.99  & 0.397 $\pm$ 0.025 \\
  & 0.7  & 33.07 $\pm$ 4.59  & 38.07 $\pm$ 16.92 & 0.640 $\pm$ 0.119 \\
  & 0.9  & 33.40 $\pm$ 7.32  & 46.20 $\pm$ 5.04  & 0.741 $\pm$ 0.038 \\
  & 0.9+ & 32.67 $\pm$ 15.97 & 42.73 $\pm$ 12.47 & 0.726 $\pm$ 0.099 \\ \hline
\multirow{4}{*}{titanic}
  & 0.3  & 23.73 $\pm$ 5.99  & 37.33 $\pm$ 2.87  & 0.399 $\pm$ 0.026 \\
  & 0.7  & 35.33 $\pm$ 3.59  & 38.53 $\pm$ 12.49 & 0.701 $\pm$ 0.126 \\
  & 0.9  & 37.87 $\pm$ 8.91  & 43.40 $\pm$ 12.24 & 0.777 $\pm$ 0.075 \\
  & 0.9+ & 34.67 $\pm$ 7.59  & 48.67 $\pm$ 10.34 & 0.781 $\pm$ 0.052 \\
\bottomrule
\end{tabular}
\caption{Baseline per-topic opinion and polarisation scores across BCM thresholds. Values reflect means and 95\% confidence intervals aggregated across model variants. The threshold 0.9+ experiments employed high-resource debunking strategies over 50 rounds with a threshold of 0.9, where the blue team generated high-potency messages during the first 20 rounds.}
\label{tab:topic-summary-baseline}
\end{table*}

\begin{table*}[t]
\centering
\footnotesize
\setlength{\tabcolsep}{4pt}
\begin{tabular}{llccc}
\toprule
\textbf{Topic} & \textbf{BCM Threshold} & \textbf{Red \% (Mean $\pm$ CI)} & \textbf{Blue \% (Mean $\pm$ CI)} & \textbf{Polarisation (Mean $\pm$ CI)} \\
\midrule
\multirow{4}{*}{alien-govt}
  & 0.3  & 22.00 $\pm$ 0.00  & 37.33 $\pm$ 5.74  & 0.395 $\pm$ 0.045 \\
  & 0.7  & 30.80 $\pm$ 10.15 & 44.33 $\pm$ 8.95  & 0.697 $\pm$ 0.126 \\
  & 0.9  & 36.20 $\pm$ 10.69 & 48.73 $\pm$ 5.45  & 0.782 $\pm$ 0.010 \\
  & 0.9+ & 32.67 $\pm$ 7.59  & 47.13 $\pm$ 9.44  & 0.792 $\pm$ 0.059 \\ \hline
\multirow{4}{*}{bird-drones}
  & 0.3  & 23.33 $\pm$ 7.59  & 35.93 $\pm$ 8.47  & 0.417 $\pm$ 0.033 \\
  & 0.7  & 34.67 $\pm$ 17.45 & 40.13 $\pm$ 18.33 & 0.693 $\pm$ 0.065 \\
  & 0.9  & 36.20 $\pm$ 0.86  & 41.33 $\pm$ 2.87  & 0.755 $\pm$ 0.052 \\
  & 0.9+ & 32.67 $\pm$ 2.87  & 47.33 $\pm$ 10.34 & 0.796 $\pm$ 0.084 \\ \hline
\multirow{4}{*}{bread-bird-lhc}
  & 0.3  & 22.00 $\pm$ 4.97  & 33.40 $\pm$ 3.02  & 0.399 $\pm$ 0.032 \\
  & 0.7  & 31.73 $\pm$ 1.15  & 44.73 $\pm$ 4.62  & 0.707 $\pm$ 0.034 \\
  & 0.9  & 38.20 $\pm$ 9.07  & 42.67 $\pm$ 5.74  & 0.758 $\pm$ 0.053 \\
  & 0.9+ & 40.60 $\pm$ 7.35  & 45.47 $\pm$ 1.15  & 0.790 $\pm$ 0.119 \\ \hline
\multirow{4}{*}{cake-moon}
  & 0.3  & 21.33 $\pm$ 2.87  & 36.00 $\pm$ 4.97  & 0.402 $\pm$ 0.029 \\
  & 0.7  & 32.40 $\pm$ 12.69 & 41.47 $\pm$ 3.19  & 0.700 $\pm$ 0.085 \\
  & 0.9  & 40.07 $\pm$ 9.69  & 42.07 $\pm$ 10.19 & 0.792 $\pm$ 0.070 \\
  & 0.9+ & 35.87 $\pm$ 5.47  & 49.60 $\pm$ 4.55  & 0.790 $\pm$ 0.106 \\ \hline
\multirow{4}{*}{chemtrails}
  & 0.3  & 21.80 $\pm$ 13.45 & 35.80 $\pm$ 4.24  & 0.382 $\pm$ 0.097 \\
  & 0.7  & 33.53 $\pm$ 5.76  & 40.00 $\pm$ 8.61  & 0.677 $\pm$ 0.050 \\
  & 0.9  & 34.67 $\pm$ 10.34 & 45.33 $\pm$ 2.87  & 0.767 $\pm$ 0.138 \\
  & 0.9+ & 39.27 $\pm$ 12.27 & 45.33 $\pm$ 11.74 & 0.814 $\pm$ 0.080 \\ \hline
\multirow{4}{*}{flat-earth}
  & 0.3  & 21.73 $\pm$ 9.75  & 35.33 $\pm$ 2.87  & 0.381 $\pm$ 0.076 \\
  & 0.7  & 30.00 $\pm$ 4.97  & 45.07 $\pm$ 9.23  & 0.670 $\pm$ 0.044 \\
  & 0.9  & 35.33 $\pm$ 5.74  & 46.00 $\pm$ 4.97  & 0.766 $\pm$ 0.065 \\
  & 0.9+ & 35.33 $\pm$ 10.34 & 48.00 $\pm$ 14.90 & 0.791 $\pm$ 0.038 \\ \hline
\multirow{4}{*}{haarp}
  & 0.3  & 21.67 $\pm$ 9.61  & 36.00 $\pm$ 0.00  & 0.407 $\pm$ 0.034 \\
  & 0.7  & 32.27 $\pm$ 20.37 & 43.47 $\pm$ 10.90 & 0.681 $\pm$ 0.113 \\
  & 0.9  & 36.67 $\pm$ 2.87  & 43.33 $\pm$ 5.74  & 0.778 $\pm$ 0.021 \\
  & 0.9+ & 36.00 $\pm$ 9.94  & 44.00 $\pm$ 8.61  & 0.775 $\pm$ 0.071 \\ \hline
\multirow{4}{*}{moon-alien-satellite}
  & 0.3  & 20.00 $\pm$ 8.61  & 37.13 $\pm$ 4.88  & 0.395 $\pm$ 0.053 \\
  & 0.7  & 34.67 $\pm$ 12.34 & 34.67 $\pm$ 10.34 & 0.643 $\pm$ 0.044 \\
  & 0.9  & 32.67 $\pm$ 7.59  & 41.33 $\pm$ 2.87  & 0.753 $\pm$ 0.029 \\
  & 0.9+ & 33.80 $\pm$ 12.89 & 46.07 $\pm$ 4.72  & 0.757 $\pm$ 0.188 \\ \hline
\multirow{4}{*}{moon-landing}
  & 0.3  & 23.67 $\pm$ 12.75 & 35.67 $\pm$ 3.98  & 0.460 $\pm$ 0.254 \\
  & 0.7  & 31.33 $\pm$ 5.74  & 42.67 $\pm$ 2.87  & 0.686 $\pm$ 0.075 \\
  & 0.9  & 38.67 $\pm$ 5.74  & 44.53 $\pm$ 15.08 & 0.768 $\pm$ 0.018 \\
  & 0.9+ & 34.07 $\pm$ 5.22  & 47.00 $\pm$ 2.48  & 0.768 $\pm$ 0.036 \\ \hline
\multirow{4}{*}{titanic}
  & 0.3  & 26.73 $\pm$ 7.37  & 31.73 $\pm$ 13.92 & 0.404 $\pm$ 0.102 \\
  & 0.7  & 32.47 $\pm$ 4.02  & 43.73 $\pm$ 1.15  & 0.691 $\pm$ 0.064 \\
  & 0.9  & 38.00 $\pm$ 4.97  & 44.67 $\pm$ 5.74  & 0.777 $\pm$ 0.032 \\
  & 0.9+ & 36.67 $\pm$ 7.59  & 42.67 $\pm$ 7.59  & 0.763 $\pm$ 0.125 \\
\bottomrule
\end{tabular}
\caption{Per-topic opinion and polarisation scores for our (resource-constrained) experiments across BCM thresholds. Values reflect means and 95\% confidence intervals aggregated across model variants. The threshold 0.9+ experiments employed high-resource debunking strategies over 50 rounds with a threshold of 0.9, where the blue team generated high-potency messages during the first 20 rounds.}
\label{tab:topic-summary-ours}
\end{table*}

\end{document}